\documentclass[letterpaper, 10 pt, conference]{ieeeconf}  

\IEEEoverridecommandlockouts                              

\usepackage{booktabs}
\usepackage{graphicx}
\usepackage{array}
\usepackage{amsmath}
\usepackage{amsfonts}
\usepackage{xcolor}
\usepackage{colortbl}

\usepackage{float}
\usepackage{placeins}
\usepackage{pifont}

\newcommand{\cmark}{\ding{51}} 
\newcommand{\xmark}{\ding{55}} %

\usepackage[compatibility=false]{caption}
\newcommand{\projname}{\textbf{ContactDP}}

\title{\textbf{ContactDP}: Contact-Guided Diffusion Policy for \\Tight Insertion Tasks
}

\author{Chengyi Xing$^{1}$, Shaoxiong Yao${^1}$, Diego Romeres$^{1}$, and Devesh K. Jha$^{1}$
\thanks{$^{1}$This work was done when the authors were at Mitsubishi Electric Research Labs, Cambridge, MA, US.
{\tt\footnotesize chengyix@stanford.edu, shaoxiong.work@gmail.com, diego.romeres@gmail.com, devesh.dkj@gmail.com}}%
}
\begin{document}

\maketitle
\thispagestyle{empty}
\pagestyle{empty}


\begin{abstract}
    High-precision connector insertion remains challenging for robotic systems due to tight mechanical tolerances, partial observability during contact, and multimodal uncertainty arising from occlusion and contact ambiguity. Successful insertion requires closed-loop contact guidance that continuously integrates global alignment cues with local contact feedback to produce stable corrective actions under interaction. In this work, we present \projname\ (Contact-Guided Diffusion Policy for Tight Insertion Tasks), a multimodal diffusion-policy framework for contact-rich insertion. \projname\ jointly integrates wrist RGB observations, fingertip tactile sensing, and wrist-mounted force–torque measurements to infer contact state and generate temporally consistent corrective motions during insertion. To ensure stable execution under contact, the learned policy operates together with a hybrid position–force controller that provides compliant low-level interaction. We evaluate our approach on a suite of industrial-grade connector insertion tasks with varying connector geometries, grasp conditions, and initial misalignment. Across all tasks, \projname\ significantly outperforms vision-only diffusion policies for performance, reliability and generalization. 
\end{abstract}



\section{Introduction}
\begin{figure}[t]
    \centering
    \includegraphics[width=0.9\linewidth]{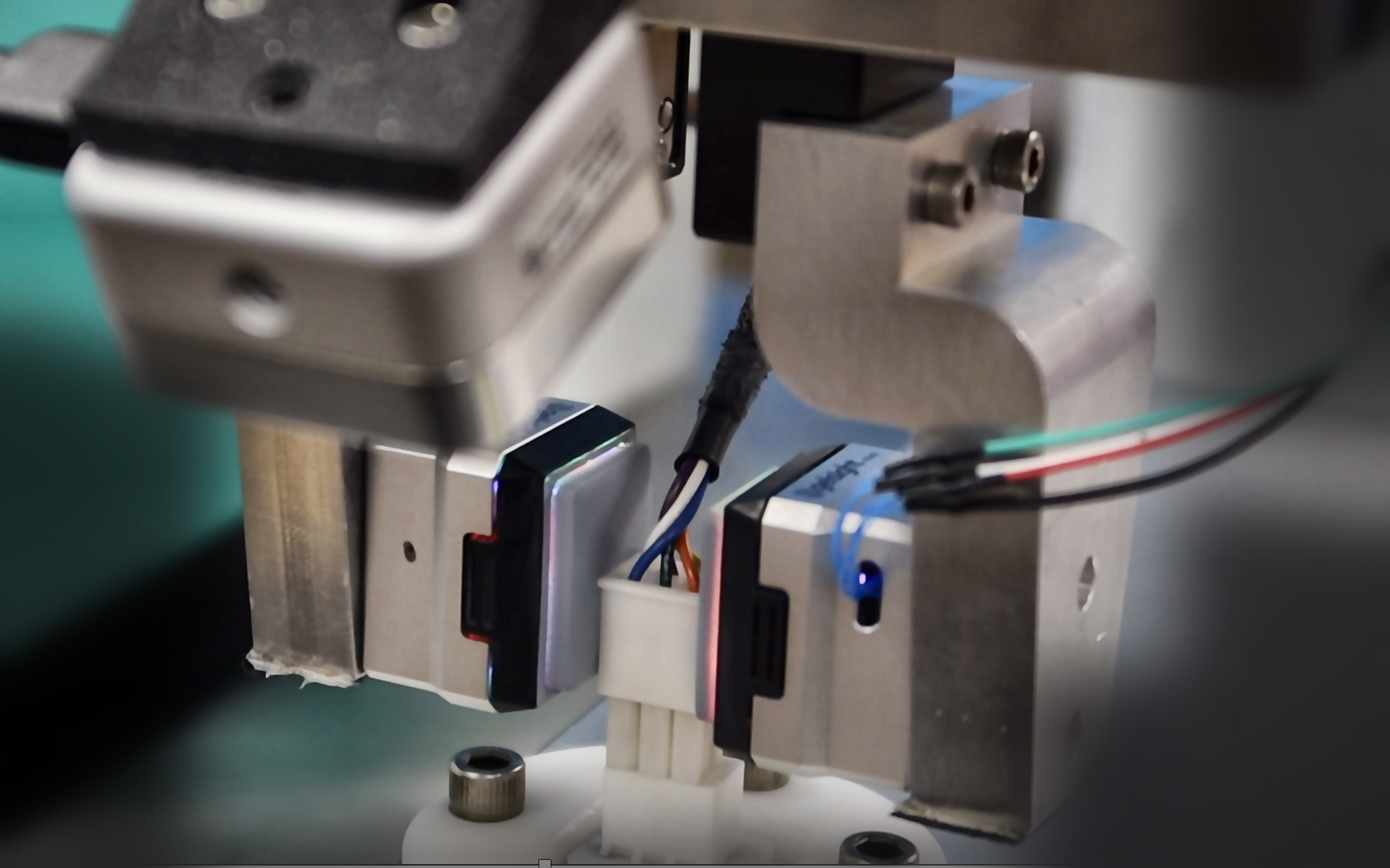}
    \caption{Due to tight tolerances, the connector insertion leads to complex contact formations in the presence of pose uncertainty. Vision-alone is often insufficient to resolve uncertainty while working with these parts. Contact-DP makes use of the additional information from force and touch to learn robust and high performance policies.}
    \label{fig:cover}
\end{figure}

Robotic insertion under partial observability requires contact and force reasoning for performance and robustness. For example, consider the task of connector insertion with industrial grade connectors shown in Figure~\ref{fig:cover}. Tasks such as connector mating, socket insertion, and tight-tolerance part seating require robots to execute precise contact interactions under very strict contact constraints. Even small pose errors can result in jamming, misaligned or tilted engagement, and excessive interaction forces that may damage components. Given the high precision required, vision alone is insufficient for reliable task completion. Force feedback and contact-guided control can be valuable for these tasks, designing a robust feedback loop that satisfies such stringent precision requirement remains mostly an open challenge. This is particularly challenging for insertion and assembly of small parts with tight tolerances (see Figure~\ref{fig:cover}) where vision-alone methods tend to struggle to achieve very high accuracy. We present a contact-guided diffusion policy for these tight tolerance, small-sized industrial grade connectors which leverage additional information by force and touch and a hierarchical controller to achieve high performance. 


Humans solve these kind of tasks by continuously integrating multiple sensing modalities during manipulation. Prior to contact, visual feedback provides global alignment cues that guide coarse positioning. Once interaction begins, however, the task becomes increasingly dominated by local contact signals such as force transients, tactile deformation, and subtle changes in contact geometry. These signals enable humans to infer insertion state, detect misalignment, and execute fine corrective motions under partial observability. Importantly, the relative utility of different sensing modalities changes throughout the task, requiring continuous reasoning over multimodal feedback.

This creates a fundamental challenge for robotic policy learning. During insertion, multiple contact configurations may produce similar observations while requiring different corrective actions~\cite{dong2021icra, 10178229}. Connector symmetries, occlusion during contact, and uncertainty in frictional interaction further increase the ambiguity. Moreover, successful insertion requires temporally consistent corrective behavior: small unstable motions can accumulate over time and lead to failure or unsafe contact forces. Policies that rely primarily on vision or deterministic action prediction therefore often struggle in tight-tolerance, contact-rich settings.

High-resolution tactile sensors enable robots to directly observe local contact geometry and surface deformation during interaction, while wrist-mounted force–torque sensing provides complementary information about contact evolution and changes in contact state. At the same time, diffusion policies offer an attractive framework to learn multimodal action distributions and generate temporally coherent corrective behaviors under uncertainty. However, integrating these modalities effectively for robust insertion remains difficult, particularly for tasks that require high precision and stable compliant interaction.

In this work, we present \projname, a multimodal diffusion-policy framework for the insertion of tight-tolerance connectors. Our central hypothesis is that robust insertion emerges from multimodal contact guidance, where vision, tactile sensing, and force feedback provide complementary information at different phases of insertion. To ensure safe and stable execution under contact, the learned policy operates together with a hybrid position–force controller that provides compliant low-level interaction.
We evaluate our approach on a suite of industrial connector insertion tasks spanning varying connector geometries, grasp conditions, and initial alignment uncertainty. Our experiments demonstrate that multimodal contact guidance substantially improves insertion robustness compared to vision-only diffusion policies and detection-based alignment baselines. Through ablation studies, we analyze the complementary role of vision, tactile sensing, and force feedback during the insertion task.

In summary, this work makes the following contributions:
\begin{enumerate}
    \item A multimodal diffusion-policy framework for insertion of tight-tolerance connectors that jointly integrates vision, tactile sensing, and force–torque feedback for contact-guided manipulation.
    \item A hybrid hierarchical control architecture that combines diffusion-based corrective action generation with compliant low-level position–force control for stable interaction under contact.
\end{enumerate}

We present extensive experimental evaluations demonstrating improved robustness, safer interaction forces, and cross-connector transfer compared to vision-only and regression-based baselines.
\section{Literature Review}
\subsection{Tactile sensing for contact-rich insertion.}
High-resolution tactile sensors such as GelSight enable robots to perceive detailed local contact geometry that vision alone cannot capture~\cite{yuan2017gelsight,lambeta2020digit}.
These tactile sensors provide dense surface deformation cues that allow the robot to infer contact-state of manipulated object~\cite{ma2021icra,kim2022icra}. 
These perception cues have enabled tactile-driven insertion strategies under significant uncertainty~\cite{dong2021icra}. 
Some methods leverage tactile to estimate accurate object pose and have grasp variations in connector-like tasks~\cite{okumura2022tactileSensitiveNewtonianvae, chang2024iros}. 
Fu et al.~\cite{fu2023icra} proposed a two-stage insertion method that manually specifies tactile sensing for alignment and vision for insertion.
These results underscore that tactile feedback is often indispensable for tight-tolerance insertion, but also reveal the difficulty of achieving robust generalization across diverse contact conditions without additional context and guidance.

\subsection{Diffusion policies for compliant contact-rich manipulation.}
Diffusion-based imitation learning has recently emerged as a powerful paradigm for learning expressive visuomotor policies from demonstrations~\cite{chi2023diffusionpolicy}. In contact-rich settings, diffusion policies are particularly attractive because they can represent multimodal action distributions and support receding-horizon rollouts that stabilize execution~\cite{lbmtri2025}. 
For tasks where visual information alone is insufficient for reliable execution, recent works have extended diffusion policies to incorporate compliance and force-aware behavior~\cite{liu2025icra}. 
For example, some approaches explicitly model compliant control variables (e.g., stiffness) ~\cite{hou2025adaptive,xu2025compliant,chen2025dexforce} or by introducing slow--fast architectures that respond quickly to tactile feedback while retaining long-horizon structure~\cite{xue2025reactiveDiffusionPolicy}. 
Our approach aligns with this direction but targets a more challenging industrial connector insertion setting with a \emph{multimodal} action-selection policy that integrates fingertip tactile images, wrist force--torque, and vision, while relying on a hybrid compliance controller for stable under-contact execution. We believe that such an architecture would unlock a large set of contact-rich, fine manipulation tasks.

\subsection{Vision--language--action models and tactile extensions.}
Generalist robot policies conditioned on language or goal specifications have made rapid progress, demonstrating broad transfer across tasks and embodiments \cite{octo2024octo,black2024pi0,pmlr-v305-black25a}. 
However, contact-rich manipulation remains challenging for VLA systems because the most informative signals often arise at the moment of contact and are not visible in RGB streams. Very recent work has begun to explicitly integrate tactile feedback into VLA pipelines by incorporating tactile features as an additional modality~\cite{yu2025forcevla}.
Beyond fusing tactile information as latent features, Tactile-VLA further integrates hybrid position–force control to enable grounded interaction~\cite{huang2025tactilevla}. VLA-Touch integrates touch in a modular manner at both the planning and control levels, refining VLA-generated actions using tactile feedback~\cite{bi2025vlatouch}.
OmniVTLA proposes a unified VTLA architecture and tactile representations designed to align tactile semantics with vision and language~\cite{cheng2025omnivtla}. However, most of these tasks treat force and/or tactile as an additional modality for task completion rather than using contact for closed-loop guidance.

To address this limitation, we propose a multimodal diffusion policy that leverages contact guidance through vision, tactile sensing, and force feedback to enable robust connector insertion under partial observability.
\section{\projname}


\begin{figure*}[htbp]
  \centering
  \includegraphics[width=.85\textwidth]{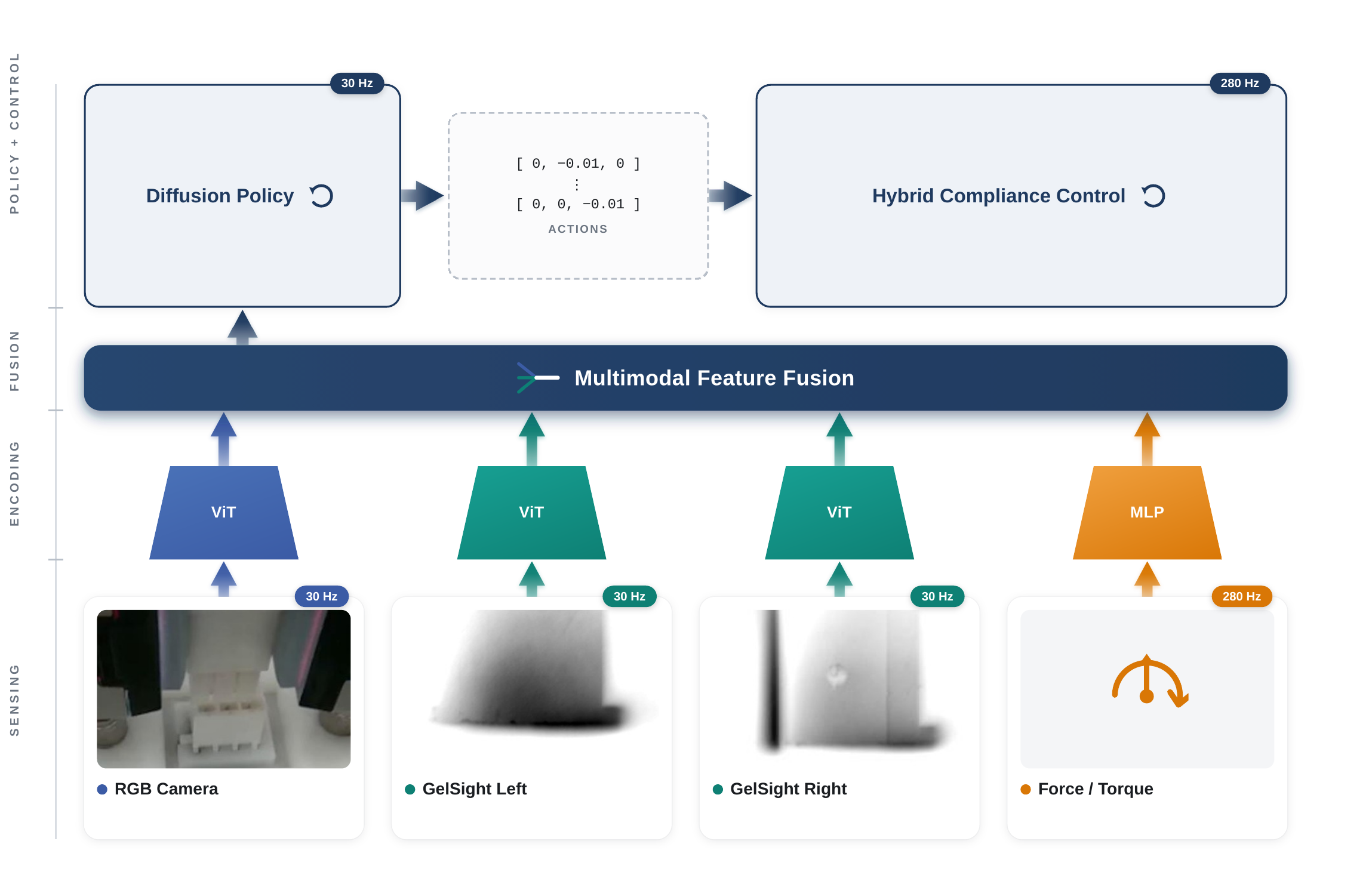}
  \caption{\textbf{System overview.} Multimodal sensory inputs are encoded and fused into a unified latent representation. A diffusion-based policy generates action trajectories, which are executed through a hybrid position–force controller for contact-rich insertion.}
\vspace{-15pt}
  \label{fig:method}
\end{figure*}

 To achieve high precision connector insertion, \projname\  incorporates multi-modal perception, diffusion policy action model, and low-level compliance control. In the following section, we first introduce our industrial insertion problem. Then we introduce the multi-modal fusion, model structure, description of our system and data collection. 

\subsection{Problem Statement}\label{subsec:problemstatement}

We study the problem of high-precision connector mating under pose uncertainty. The objective is to design a single policy that can successfully mate a diverse set of connectors, despite tight mechanical tolerances, partial observability, and geometric symmetries, using multi-modal sensory feedback. These connectors exhibit varying cross-sections, asymmetric and symmetric features, and clearance on the order of tens of microns, making the task highly sensitive to small pose errors.

An episode is considered successful if the policy reaches a desired goal set $\mathcal{G}^i$ within a finite time interval. Execution is terminated and labeled as failure if the end-effector violates workspace safety constraints,
\[
|x(t) - x^i_*| > \Delta_{\mathrm{safety}} \quad \text{or} \quad |y(t) - y^i_*| > \Delta_{\mathrm{safety}},
\]
or if a time limit $t > t_{\mathrm{max}}$ is exceeded. These constraints reflect collision avoidance, hardware safety, and practical deployment considerations.

\subsection{Action Model}
\label{subsec:diffusionpolicy}
The connector mating task requires resolving geometric ambiguity, contact uncertainty, and connector symmetries under partial observability, while producing temporally consistent corrective motions. To address these challenges, we employ a diffusion-based policy that generates action trajectories conditioned on multi-modal observations. Generating trajectory segments, rather than single-step actions, enables the policy to reason over contact evolution and recover from transient misalignment during insertion.

\paragraph*{Action Representation}
At each control timestep $t$, the policy outputs a sequence of Cartesian actions
\[
A_{t:t+H-1} = \{a_t, a_{t+1}, \dots, a_{t+H-1}\}, 
\quad a_t \in \mathbb{R}^3,
\]
where each action specifies an incremental update to a virtual end-effector target along the $x$, $y$, and $z$ axes. Predicting a trajectory allows the policy to generate smooth and temporally coherent motions, which are critical for stable insertion under tight mechanical tolerances.

\paragraph*{Multi-modal Observations and Conditioning}
The policy is conditioned on the history of multi-modal observations $o_{1:t}$. 
The GelSight tactile image $o^g_t$ captures local contact geometry and surface deformation during insertion. To emphasize contact-induced changes and suppress static bias, we preprocess the tactile signal by subtracting the initial frame,
\[
o'_g(t) = o_g(t) - o_g(0),
\]
and encode the resulting signal using a ResNet backbone. Because industrial connectors often exhibit flat or weakly textured surfaces, tactile feedback alone may be insufficient to infer corrective directions during early alignment. Therefore, the wrist-mounted RGB image $o^w_t$ provides global alignment cues prior to contact.  We crop $o^w_t$ to the connector region and encode it using a Vision Transformer to obtain visual features.

Force-torque measurements $o^f_t$ provide direct feedback on contact forces and moments, enabling the policy to infer contact state, insertion depth, and impending failure. 
Features from all modalities are fused into a unified latent representation, which conditions the diffusion-based decision module at each timestep.

\paragraph*{Diffusion-Based Policy}
Conditioned on the fused observation features, the policy models the conditional distribution
\[
\pi_\theta(A_{t:t+H-1} \mid o_{1:t})
\]
using a diffusion process. Due to connector symmetries and partial observability, multiple action trajectories may be valid from the same observation history, particularly during early alignment. Diffusion policies naturally capture such multi-modal action distributions while maintaining temporal consistency across the predicted trajectory.

At inference time, the policy starts from a Gaussian noise sample and iteratively refines a noisy action trajectory through a fixed number of denoising steps. The denoising network is conditioned on the fused multi-modal representation at each step, allowing the policy to adapt its actions based on both global alignment cues and local contact feedback.

\paragraph*{Training Objective.}
The policy is trained using the standard denoising diffusion objective. Given a ground-truth action trajectory $A_{t:t+H-1}^{(0)}$, Gaussian noise is added according to a predefined noise schedule to obtain a noisy trajectory $A_{t:t+H-1}^{(k)}$ at diffusion step $k$. The denoising network is trained to predict the injected noise conditioned on the observation history $o_{1:t}$. The training loss is defined as the mean squared error between the predicted and true noise.

Although the training objective is standard, its application in this setting enables the policy to learn a robust conditional action distribution that accounts for sensor noise, contact variability, and connector-specific geometric constraints. A single diffusion model is shared across all connectors, allowing the policy to generalize across insertion tasks without connector-specific heads or retraining.

\subsection{Hybrid position--force controller.}
We employ a low-level hybrid position--force controller to ensure smooth insertion and prevent excessive contact forces that could damage the connector or socket. The controller is implemented as an axis-dependent impedance controller with different stiffness parameters along orthogonal axes. Specifically, the robot operates with high stiffness in the lateral ($x$–$y$) plane, effectively enforcing position control to enable precise alignment with the socket. Along the insertion ($z$) axis, the controller uses a low stiffness setting, allowing compliant motion that regulates contact forces during insertion. This anisotropic stiffness profile enables the robot to correct lateral misalignment while safely advancing along the insertion direction under contact.

The high-level diffusion policy operates asynchronously with respect to the low-level controller and outputs incremental updates to a virtual end-effector target. The low-level controller tracks this target at a higher control frequency, executing small motion steps to ensure stable and smooth interaction throughout contact-rich insertion.

\subsection{Data Collection}\label{subsec:datacollection}

To collect demonstration data for contact-rich insertion while minimizing manual effort, we design an automatic data collection pipeline that executes a structured insertion procedure and records synchronized observations and actions. 

Each demonstration is segmented into three semantically meaningful phases that reflect the physical structure of the insertion task:
\begin{itemize}
    \item Approach: moving toward the socket until initial contact is established,
    \item Alignment: correcting lateral misalignment under contact,
    \item Insertion: advancing the connector fully into the socket.
\end{itemize}
These phases are determined automatically based on force and position signals during execution and are used for data annotation.


\begin{figure}[t]
    \centering
    \includegraphics[width=0.8\linewidth]{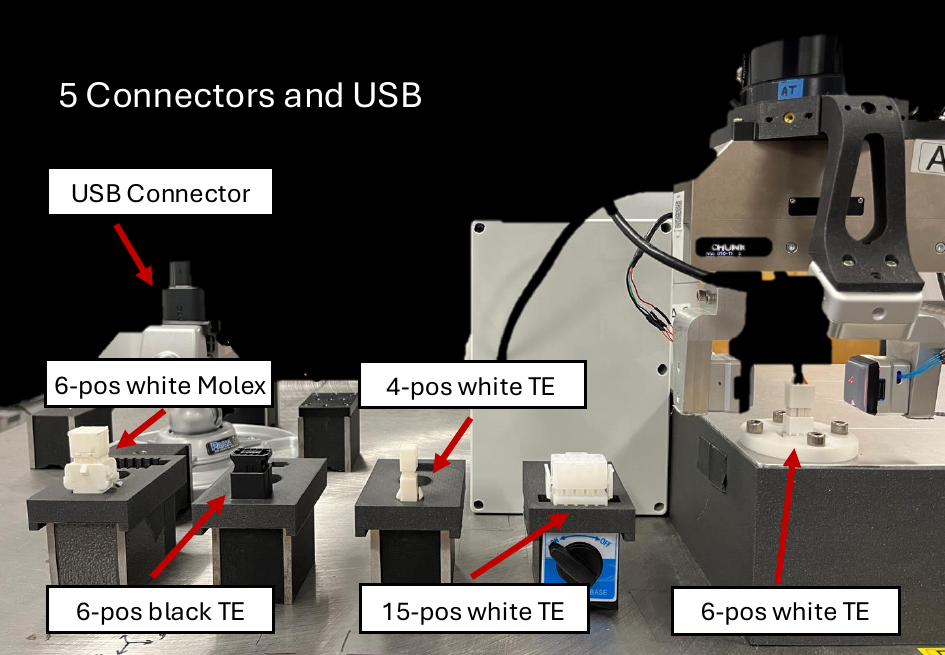}
    \caption{Experiment setup with the 6 connectors used for the study. We introduce known errors in the initial location and grasp when the robot makes an insertion attempt. During training, that provides us labels and during test, we use it to evaluate the generalization.}
    \label{fig:connectors}
    \vspace{-20pt}
\end{figure}

\paragraph*{Automatic demonstration generation}
Given the nominal fully aligned pose $p^i_*$ for connector $c_i$, the robot is first moved to a randomized initial position $p^i$ as defined in Sec.~\ref{subsec:problemstatement}. From this initial state, a compliance controller executes the insertion procedure while all observations and actions are recorded. 
During the \emph{approach} phase, the end effector is commanded to move  along the insertion axis until contact with the socket is detected. Contact is identified when the measured normal force exceeds a predefined threshold, $F_z > z_{th}^1.$ Once contact is established, the controller transitions to the \emph{alignment} phase. We use a hybrid force position controller during this phase where the insertion axis is controlled in force and the lateral axis are controlled in position using a simple proportional law to align with the insertion location. Lateral alignment is achieved using a proportional controller in the horizontal directions:
\[
A_x = -k_P (x_r - x^i), \quad
A_y = -k_P (y_r - y^i),
\]
where $(x_r, y_r)$ denotes the current end-effector position and $(x^i, y^i)$ the nominal aligned position for connector $c_i$. The proportional gain $k_P$ is chosen appropriately to minimize chatter around the goal. 
The alignment phase terminates when the measured normal force drops below a second threshold $z_{th}^2$.

Finally, during the \emph{insertion} phase, the connector is advanced along the insertion axis until successful mating is achieved, defined by $z_r < z_{th}^i.$ The above procedure for data collection makes use of contact force guidance for automatic alignment detection and data collection.

\section{Experiments and Results}

\begin{table*}[t]
\centering
\small
\begin{tabular}{lcccc|cccccc}
\hline
Connector &
Vision &
Force &
Tactile &
$S$ (succ/eps) &
$\bar{T}_s \; [\mathrm{s}]$ &
$\bar{F}_z$ (all) &
$\bar{F}_z$ (success) &
$F_z^{\max}$ (all) &
$F_z^{\max}$ (success) \\
\hline
6-pos      & \cmark   &  \xmark  &  \xmark        & 0.657 & 39.696 & 4.4052 & 4.1888 & 12.2156 & 12.1275 \\
15-pos      & \cmark   &  \xmark   &  \xmark      & 0.704 & 25.097 & 0.7136 & 0.7096 & 2.5376 & 2.4724 \\
4-pos      & \cmark    &   \xmark   &  \xmark    &0.400 & 24.291 & 5.6305 & 1.1236 & 13.5676 & 3.6785 \\
\hline
6-pos      & \cmark   &  \cmark   &  \xmark      & 0.874 & 25.488 & 1.8495 & 1.9837 & 4.4144 & 4.7361 \\
15-pos      & \cmark   &  \cmark   &  \xmark      & 0.932 & 21.123 & 0.4037 & 0.4143 & 1.3513 & 1.3852 \\
4-pos      & \cmark    &   \cmark   &  \xmark    &0.890 & 29.932 & 2.4673 & 2.7333 & 6.2865 & 6.9468 \\
\hline
6-pos           & \cmark   & \cmark & \cmark & \textbf{0.977} & 27.402 & 2.3926 & 2.4186 & 5.1544 & 5.2360 \\
15-pos           & \cmark   & \cmark & \cmark & \textbf{1.000} & 21.755 & 1.3385 & 1.3385 & 3.5722 & 3.5722 \\
4-pos             & \cmark  & \cmark     & \cmark  & \textbf{0.958} & 28.739 & 1.9020 & 1.7425 & 5.2633 & 4.9548 \\
\hline
\end{tabular}
\caption{
%
%
ContactDP sensing-modality ablation. Cumulatively adding wrist force–torque, then fingertip GelSight tactile sensing, to a vision-only policy. Force feedback raises success rate and improves the other metrics; tactile sensing improves them further, evidence that tactile adds value beyond force, likely by conveying additional contact-pose information.}
\label{tab:grasp_results}
\end{table*}

We evaluate our approach on a suite of industrial connector insertion tasks designed to reflect real-world assembly conditions, including tight mechanical tolerances, limited visual observability during contact, and connector-specific geometry. 
The experiments are conducted on a 6-DoF industrial manipulator equipped with a GelSight Mini tactile sensor, a wrist-mounted force–torque sensor, and an RGB wrist camera (Fig. \ref{fig:connectors}). 
Through our experiments, we try to understand the performance of the proposed method against a baseline vision-alone diffusion policy. 
We further evaluate the zero-shot and few-shot generalization capabilities of the proposed method to previously unseen connectors.


\subsection{Experiment and benchmark system setup}

\begin{figure}[hb]
    \centering
    \includegraphics[width=\linewidth]{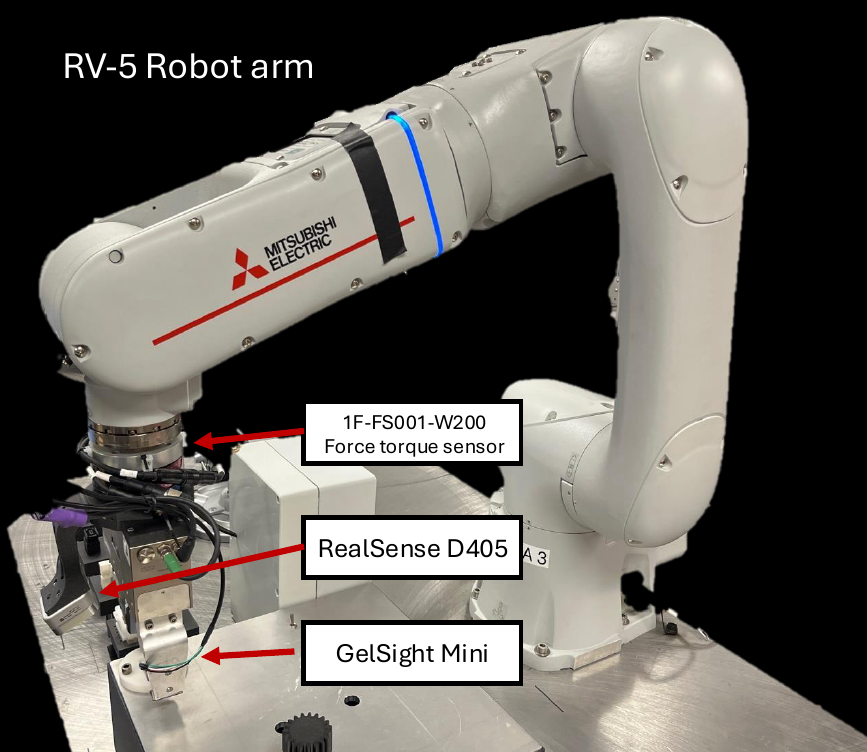}
    \caption{Experimental robotic manipulation setup. A Mitsubishi Electric RV-5 industrial robot arm is equipped with a wrist-mounted 6-axis force–torque sensor (1F-FS001-W200), an Intel RealSense D405 RGB-D camera, and a GelSight Mini tactile sensor at the end effector. }
    \label{fig:setup}
    \vspace{-10pt}
\end{figure}
\begin{table*}[t]
\centering

\begin{tabular}{lcccccc}
\hline
Dataset & $S$ (succ/eps) & $\bar{T}_s \; [\mathrm{s}]$ & $\bar{F}_z$ (all) & $\bar{F}_z$ (success) & $F_z^{\max}$ (all) & $F_z^{\max}$ (success) \\
\hline
Vision-based DP & 0.587 & 29.67 $\pm$ 7.07 & 3.58 $\pm$ 2.09 & 2.00 $\pm$ 1.55 & 9.43 $\pm$ 4.91 & 6.09 $\pm$ 4.29 \\
\textbf{\projname\ (Ours)} & \textbf{0.978} & 25.97 $\pm$ 3.03 & 1.88 $\pm$ 0.43 & 1.83 $\pm$ 0.45 & 4.66 $\pm$ 0.77 & 4.59 $\pm$ 0.73 \\
\hline
\end{tabular}
\caption{Overall insertion performance on the in-domain connector benchmarks. 
We compare vision-based diffusion policy (DP) and our \projname\ using success rate $S$, mean completion time $\bar{T}_s$ over successful episodes, and mean/peak vertical interaction force ($\bar{F}_z$, $F_z^{\max}$) reported over all episodes and successful episodes only. Each connector is tested over $50$ episodes.
}
\label{tab:comparison}
\end{table*}

We conduct all experiments using a Mitsubishi Electric MELFA Assista RV-5AS-D industrial robot arm with a repeatability of $0.03,\mathrm{mm}$. 
A six-axis force–torque sensor (Mitsubishi Electric 1F-FS001-W200) is mounted between the robot flange and the gripper to measure interaction forces during insertion. 
The end effector is equipped with a Schunk WSG50 parallel gripper, with two GelSight Mini tactile sensors mounted on the gripper fingers. 
An Intel RealSense D405 RGB camera is rigidly mounted at the wrist, providing a fixed viewpoint relative to the end effector. 
The model runs at $30\,\mathrm{FPS}$ on a computer equipped with an NVIDIA RTX~4090 GPU. 


Figure~\ref{fig:connectors} illustrates the six connectors used in our experiments. According to their technical specifications, all linear dimensions are manufactured with a tolerance of $\pm 0.13\,\mathrm{mm}$, and all angular dimensions with a tolerance of $\pm 2^\circ$. 
As also evident from the qualitative visualizations in Figure~\ref{fig:qual}, the resulting clearances are visually minimal, underscoring the tight alignment required for successful insertion. 
Due to their asymmetric and polarized geometry, the connectors admit only a single valid insertion orientation. Even small pose errors therefore result in mechanical interference, making precise alignment under contact essential.

Three connectors are treated as in-domain and are used for both training and evaluation:
\begin{itemize}
    \item \textbf{6-pos white TE connector}: a compact multi-pin connector with narrow insertion clearances and limited contact surface area, requiring precise lateral alignment. Specifically, the 6-pos Mini-Universal MATE-N-LOK™ plug housing (PN~172168-1) and receptacle housing (PN~172160-1), TE Connectivity.
    \item \textbf{15-pos white TE connector}: a connector with the widest base and guiding features, offering a more forgiving geometry. Specifically, the 15-pos Universal MATE-N-LOK™ plug housing (PN~1-480710-0) and receptacle housing (PN~926647-1), TE Connectivity.
    \item \textbf{4-pos white TE connector}: a smaller-scale variant with tighter tolerances and increased sensitivity to millimeter-scale pose errors, making it the most challenging in-domain task. Specifically, the 4-pos Mini-Universal MATE-N-LOK™ plug housing (PN~172167-1) and receptacle housing (PN~172159-1), TE Connectivity.
\end{itemize}

We additionally evaluate generalization on three out-of-domain connectors not seen during training:
\begin{itemize}
    \item \textbf{6-pos black TE connector}: a 6-pos Micro MATE-N-LOK™ receptacle housing (PN~1-178127-6) and plug housing (PN~1-178137-2), TE Connectivity.
    \item \textbf{6-pos white Molex connector}: a 6-pos Molex receptacle housing (PN~0050842062) and polarized plug housing (PN~0050841065), Molex.
    \item \textbf{USB connector}: a standard USB connector.
\end{itemize}
These connectors differ in geometry, pin arrangement, and surface appearance, and are used to assess cross-connector generalization of the learned policy.

For each of the three in-domain connectors (small, mini, large), we collect 100 demonstrations as described in Sec.~\ref{subsec:datacollection}. Each demonstration starts from a randomized initial pose and proceeds until successful insertion or termination due to safety constraints. The diffusion-based policy described in Sec.~\ref{subsec:diffusionpolicy} is trained jointly on the aggregated dataset from all three connectors, yielding a single shared policy without connector-specific heads or retraining.

\subsection{Implementation Details}
We implement a diffusion policy with a planning horizon of $H=16$, conditioned on $n_{\text{obs\_steps}}=2$ observation steps and producing $n_{\text{action\_steps}}=8$ actions per inference without latency steps. 
The diffusion process uses a DDIM noise scheduler with 100 training timesteps and 2 inference steps. 
Visual observations are encoded with a ViT-B/16 backbone, with ImageNet normalization and cropping to focus on the connector. 
Training is performed with a batch size of 64 for 10 epochs, and an exponential moving average (EMA) of model parameters is maintained throughout training to stabilize learning.

\subsection{Baselines}
We compare the proposed \projname\ against the following baselines across multiple tasks.


\paragraph{Vision-based Diffusion Policy}
We consider a strong vision-only diffusion policy~\cite{chi2023diffusionpolicy} trained on the same hardware platform, task distribution, and demonstration sequences as our method. 
This baseline employs a similarly high-capacity imitation learning architecture while relying exclusively on visual observations.

\subsection{Main results}
\begin{figure}[t]
  \centering
  \includegraphics[width=0.48\textwidth]{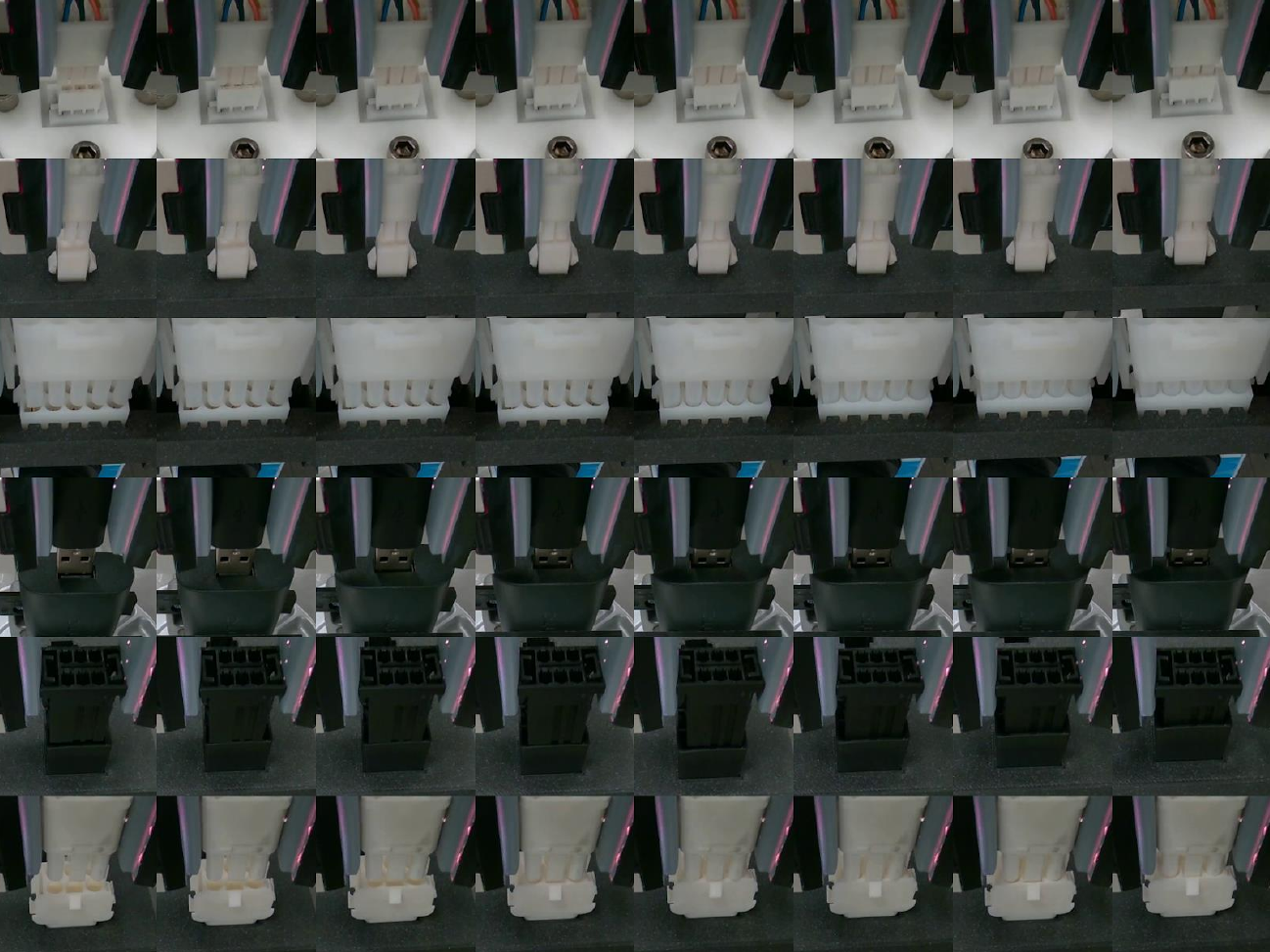}
  \caption{Qualitative results across six connector insertion tasks. From top to bottom, the rows show: (1) 6-pos white TE connector, (2) 15-pos white TE connector, (3) 4-pos white TE connector, (4) a standard USB 3.0 connector, (5) 6-pos black TE connector, and (6) 6-pos white Molex connector.}
\label{fig:qual}
\vspace{-15pt}
\end{figure}

At test time, the learned policy is evaluated in closed loop on each connector. 
To quantify performance, we report the following metrics:
\begin{itemize}
    \item Success rate, $S = \frac{N_{\text{succ}}}{N_{\text{eps}}}$,
    where $N_{\text{succ}}$ is the number of successful insertions and $N_{\text{eps}}$ is the total number of evaluation episodes. We use $50$ episodes per connector to generate the evaluation metrics.

    \item Mean completion time (successful episodes only), $\bar{T}_s \; [\mathrm{s}]$, computed over episodes that end in a successful insertion.

    \item Mean vertical interaction force, denoted by $\bar{F}_z$ (all episodes) and $\bar{F}_{z,\text{succ}}$ (successful episodes), is defined as the time-averaged force along the $z$ axis over an episode. 
    These metrics capture the typical contact load during deployment and under successful insertions, respectively.



    \item Peak vertical interaction force, denoted by $F_z^{\max}$ (all episodes) and $F_{z,\text{succ}}^{\max}$ (successful episodes), is defined as the maximum force along the z axis observed within an episode. 
    These metrics characterize the worst-case contact load during deployment and under successful operation, respectively.


\end{itemize}

Force statistics are reported both over \emph{all episodes} and over \emph{successful episodes only}, enabling separation of contact quality from failure handling.

At test time, the learned policy is evaluated in closed loop on each connector. 
Table~\ref{tab:comparison} summarizes the evaluation across baselines and our method. 
Our method achieves a success rate of $S=0.978$, demonstrating near-deterministic insertion performance under the same evaluation protocol. 
This gain is not obtained by applying excessive force. Across all episodes, the mean vertical interaction force remains moderate ($\bar{F}_z=1.878$), and when conditioned on successful trials it further decreases to $\bar{F}_z(\text{success})=1.833$. 
These results indicate that the learned policy consistently converges to low-load insertion trajectories rather than brute-forcing through contact.
In contrast, the baseline vision-only Diffusion policy achieves a success rate of $0.587$ averaged over the three connectors. It is noted that the vision-alone DP also makes use of the hybrid position-force controller during implementation. This helps keeps the interactive forces during insertion attempts close to the proposed method. 
The main difference compared to the proposed method is the vision-alone DP does not have the force/touch based reasoning during insertion.

\begin{table*}[ht]
\centering
\small
\begin{tabular}{ccccccc}
\hline
Demos &
$S$ &
$\bar{T}_s \; [\mathrm{s}]$ &
$\bar{F}_z$ (all) &
$\bar{F}_z$ (success) &
$F_z^{\max}$ (all) &
$F_z^{\max}$ (success) \\
\hline
0  &
0.357 &
26.761 &
1.4561 &
1.3252 &
3.9760 &
4.1022 \\
10 &
0.900 &
24.299 &
0.9018 &
0.8328 &
3.1159 &
3.0271 \\
\hline
\end{tabular}
\caption{Generalization to the unseen 6-pos white Molex object without any new demonstrations and fine-tuning with 10 demonstrations.}
\label{tab:unseen_white}
\vspace{-10pt}
\end{table*}

Peak force statistics show a similar trend.
While our policy occasionally experiences higher transient contact events, the worst-case peak over all trials remains bounded ($F_z^{\max}(\text{all})=4.663$), and is slightly lower when considering only successful insertions ($F_z^{\max}(\text{success})=4.588$).
This gap suggests that failures are not driven by extreme impulsive contacts; instead, the rare unsuccessful episodes arise from timeout rather than unsafe force application.

Finally, our method completes successful insertions in $\bar{T}_s=25.965$ seconds on average.
Overall, our method provides a substantially improved reliability--safety profile, combining high success with controlled average and peak interaction forces.

\subsection{Ablation study}
\begin{table}[t]
\centering
\begin{tabular}{lcc}
\hline
\textbf{Connector Type} &  \textbf{Vision-based DP} & \textbf{\projname} \\
\hline
6-pos black TE & 0.033 & \textbf{0.286} \\
6-pos white Molex & 0.0 & \textbf{0.357} \\
USB connector & 0.477 & \textbf{0.667} \\
\hline
\end{tabular}
\caption{Insertion success rate in the zero-shot setting across three different connector types.}
\label{tab:success_rates}
\vspace{-10pt}
\end{table}

Table~\ref{tab:grasp_results} reports performance across the 4-, 6-, and 15-position connector variants and ablates the contributions of force and tactile sensing. Two trends are particularly clear. First, adding force feedback substantially improves insertion performance over vision alone, increasing the success rate from 40.0--70.4\% to 87.4--93.2\% across the three connectors. Force feedback also generally reduces completion time and contact forces, suggesting that detecting contact helps the policy adjust its motion during insertion.

Second, tactile sensing further improves the success rate to 95.8--100\%, despite the already strong performance with vision and force. We hypothesize that the high-resolution tactile observations provide richer information about local contact geometry and the pose of the grasped connector, enabling more precise corrective actions during tight-clearance insertion. Overall, the results demonstrate the complementary benefits of force and tactile feedback: force sensing provides a large improvement over vision alone, while tactile sensing further improves the reliability of insertion.

\subsection{Generalization study}
In the zero-shot setting, the policy exhibits meaningful transfer across connector types (Table~\ref{tab:success_rates}, Figure~\ref{fig:qual}). 
The policy achieves non-trivial success on both the \emph{completelyunseen} 6-pos black TE connector ($S{=}0.286$) and the white 6-pos Molex connector ($S{=}0.357$), indicating that the learned behavior is not narrowly specialized to a single connector instance. 
Notably, the substantially higher success on the USB connector ($S{=}0.667$) suggests that generalization improves when the mating tolerances are more forgiving. 
Importantly, these results also suggest robustness to substantial appearance variation: the policy transfers across connectors with markedly different visual characteristics by leveraging a common-sense insertion routine and tactile feedback to guide alignment and mating.

Furthermore, adaptation is highly sample-efficient as shown in Table~\ref{tab:unseen_white}: with only $10$ episodes of fine-tuning on the target connector, success increases to $S{=}0.900$. Fine-tuning also improves contact quality, reducing the mean interaction force from $\bar{F}_z{=}1.46$ to $0.90\,[\mathrm{N}]$ (all episodes) and lowering peak forces from $F_z^{\max}{=}3.98$ to $3.12\,[\mathrm{N}]$, while slightly shortening completion time ($26.76 \rightarrow 24.30\,[\mathrm{s}]$). Together, these results suggest that the policy captures transferable insertion primitives, and that modest on-target adaptation is sufficient to recover high success rates while maintaining safe, controlled interaction forces.

\section{Conclusion}
\label{sec:conclusion}

Insertion of industrial-grade connectors presents unique perception and learning challenges due to the precision requirements in the presence of large uncertainties as well as the contact-rich nature of the task. These tasks require closed-loop reasoning of contact events to provide very accurate contact guidance to achieve robust performance. In this paper, we introduced \projname\ which proposes the use of contact guidance using force and touch sensing to resolve uncertainty during tight tolerance connector insertion tasks.
We propose a two-tier controller design which makes use of force and touch reasoning for uncertainty compensation as well as execution of the corrective actions during insertion. We have shown that making appropriate use of force and touch, we can achieve very high performance as well as reliability when performing tight tolerance tasks fro various kinds of connectors. In the future, we would like to scale this approach using language to label different stages of the insertion process to achieve semantic understanding and general purpose behavior. Another aspect that would be interesting to explore is the grasping as part of the insertion process so that the model can grasp a connector from any pose.



\section*{ACKNOWLEDGMENT}
Figure~\ref{fig:method} on method overview was generated using Claude's image generation tool based on author-provided descriptions and subsequently edited by the authors.

\bibliographystyle{IEEEtran}
\bibliography{references}  

\end{document}